# ReMaD: Tuning-free Domain Adaptation for Classification and Out-of-Distribution Detection

Elijah Bolluyt, Cristina Comaniciu

*Abstract*— **We introduce Reduced-rank Mahalanobis Distance (ReMaD), a novel prototypical distance-based refinement to classification and out-of-distribution (OOD) detection using pretrained models without finetuning. We use embeddings of the target dataset to fit closed-form distribution statistics in the model's latent space which can classify in-distribution samples and detect OOD samples, all without training or prior knowledge of the OOD data. Building on prototype classification and OOD detection, we analyze the distribution properties of large pretrained models when processing new datasets; based on this analysis, we formulate a simple modification to Mahalanobis Distance to adapt models' latent space distributions to new domains by removing unused features, without the finetuning or hyperparameter searches required by other adaptation procedures. We demonstrate the efficacy of this method to adapt existing large pretrained image embedding models to new classification domains outside their trained capabilities by testing across four target datasets, with competitive performance in both classification and OOD detection.**

*Impact Statement* — **Neural network image classifiers deployed in real tasks must often handle unexpected data from far outside their training sets, especially in safety-sensitive applications where an inappropriate response could cause harm. While large pretrained models have become prevalent in classification tasks, current literature on methods to handle anomalous data has mostly focused on traditional trained classifiers or models which have at least observed relevant data during training. Our method allows these large models to be more safely deployed without the high cost of retraining by enhancing their ability to identify and handle out-of-distribution data. In contrast to other methods, we also do not require any hyperparameter tuning, further minimizing deployment cost and complexity. By adding this capability without significant added cost, our method can help to fortify existing systems against unexpected inputs.**

## I. Introduction

As neural network classifiers are deployed in real-world applications, the rise of large pretrained models in both image and text modalities has reduced the need to train specialized models on individual classification tasks. These large models benefit from pretraining on largescale datasets, often performing competitively with specially trained models. In addition, distillation and related methods [1] [2] can reduce these large, general models to smaller sizes, allowing deployment to constrained environments and negating the primary disadvantage of large architectures. For tasks related to their training domain, these models can often complete tasks in a zero- or few-shot modality [3] [2]. For tasks outside their original domain, these models can be finetuned; efficient methods can make this process more affordable, but it often still requires significant computational resources and specialized techniques for individual domains.

Simultaneously, robustness to anomalous inputs from outside a classifier's target distribution has become increasingly important as models are deployed to a widening array of safety-critical tasks. Detecting and handling unexpected inputs appropriately ensures that systems perform their tasks reliably once deployed. Out-of-Distribution (OOD) detection codifies this task by augmenting classifiers to detect samples which do not belong to any known in-distribution (ID) classification label. OOD detection can be implemented as a post-hoc method on trained classifier models, or via finetuning methods.

However, most existing OOD detection methods for classifiers rely on the models' differing responses to in-distribution (ID) data versus OOD data [4] [5] [6]. Large unimodal encoders which offer rich embedding spaces must be finetuned to serve as classifiers and OOD detectors. Zero-shot classifiers based on Contrastive Language Image Pretrained (CLIP) models can be augmented with OOD detection, but most methods require finetuning [7]; others assume that the model's text encoder has knowledge of the ID sample's classes [8] [9], limiting their application to tasks within the domain of the model's training data. This also largely excludes the application of general models to tasks with highly technical terminologies, such as medical data.

Zero-shot classification and OOD detection requires that the model has knowledge of the in-distribution's classes, and finetuning requires compute resources and the storage of specialized models for every task. We propose a third approach based on the hypothesis that the embedding model's features, which are trained over largescale data, are sufficiently capable to succeed at a new classification task if we can appropriately define the target distribution. To this end, we analyze the

This work has been submitted to the IEEE for possible publication. Copyright may be transferred without notice, after which this version may no longer be accessible.

This work is supported in part by ACC-New Jersey under Contract No. W15QKN-18-D-0040

E. Bolluyt is with Stevens Institute of Technology, Hoboken, NJ, 07030, USA (e-mail: ebolluyt@stevens.edu).

C. Comaniciu is with Stevens Institute of Technology, Hoboken, NJ, 07030, USA (e-mail: ccomanic@stevens.edu).

properties of latent distributions in large embedding models when utilized for new tasks; we leverage our findings to create a simple modification to existing distance-based methods to better serve domains which are novel to the model. Specifically, we address cases where a training set is available for a new domain, but the pretrained model need not have any knowledge of it, and finetuning of parameters or hyperparameters is not performed. We also exclude prior knowledge of any OOD data, in contrast to methods which include such information [10].

We introduce a novel training-free and hyperparameter tuning-free approach for the domain adaptation of large pretrained models to new classification tasks, with accompanying OOD detection capability. Related to distance-based OOD detection using Mahalanobis Distance [11] [12] [13] and prototypical networks [14], our method calculates closed-form distribution statistics using a single inference pass across a target dataset. Once collected, these statistics enables both classification and OOD detection using large pretrained embedding models, even when the target task is outside the knowledge of the model. This process requires no training, only inference, drastically reducing the computational requirements of adapting large models to new domains. The resultant classifiers leverage the large model's learned features and possess robust OOD detection capabilities.

## II. Related Work

### *A. Distance-Based OOD Detection for Trained Classifiers*

OOD detection using Mahalanobis distance [11] assumes that a trained classifier learns to cluster samples belonging to each class within the latent space produced by the model's penultimate layer. The mean location of each class's training samples within the latent space and a tied covariance matrix of the sample's distribution around these class means define the samples' distribution in the latent space. When test samples are classified, the relative likelihood that they do not belong to any known class is calculated as the squared Mahalanobis distance from the sample's latent vector to the nearest class center.

Other performant OOD detection methods for trained classifiers have more stringent requirements which make them ill-suited to our goals. For example, Maximum Softmax Probability relies on the model's final classification logit output [4] [5], while K-Nearest Neighbors for OOD detection requires comparison to individual known samples at inference time, which can become expensive for large datasets [15].

### *B. Zero-shot classification and OOD detection methods for Contrastive Language Image Pretrained (CLIP) models*

While most large pretrained embedding models must be finetuned to function as classifiers, CLIP models' multimodal alignment enables zero-shot classification on image tasks where descriptions of the classes are within the learned domain of the model [3] [2]. However, this capability does not extend to specialized cases where the target classification task falls outside the model's experience.

Several OOD detection methods have been proposed for CLIP models [7]; some learn new tokens for the text encoder through training or employ other finetuning methods. Others can operate in a training-free setting, but only in cases where the class labels are within the model's pretraining [8] [9].

### *C. Mahalanobis Classification and Prototypical Networks*

Mahalanobis distance has been widely used for classification in classical machine learning [16]. In the context of neural networks, it has been applied to classification on latent space representations in methods related to Prototypical Networks [14], where mean representations are used as "prototypes" of the corresponding classes, and classification is performed by comparing new samples to these centroids. While the original method used Euclidean distance, improved methods in this vein can outperform baseline distance metrics on training-free domain adaptation for classification and operate predominantly in few-shot settings. However, these extensions generally require training model parameters or tuning hyperparameters to adapt to new domains [17] [18] [19]. This represents a trade-off in comparison to our method, which uses the information from more data samples to adapt to new domains with no finetuning or hyperparameter searches. Many improved prototypical methods also focus specifically on classification, whereas we prioritize high performance in OOD detection for robustness.

## III. Method

We seek to adapt large pretrained models as classifiers with robust OOD detection capability, without the need for finetuning, hyperparameter searches, or prior knowledge from training. We base our method on OOD detection using Mahalanobis distance; although it was introduced for trained classifiers, this method is fully post-hoc and does not require specialized training modifications. Additionally, it uses only the latent space and not the classification logits of trained models, unlike many widely used OOD detection methods [4] [5] [20]. However, the OOD method was still formulated for trained classifiers, and its core mathematical assumption of clustering within the latent space is predicated upon the model being trained as such [11]. We will show that this method can be adapted to our use case through low-cost modifications.

### *A. Statistic Collection through Inference*

This method begins by gathering two statistics describing the training data's distribution in the model's latent space: class mean vectors, and a tied covariance matrix. Using a single inference pass of the training set, we collect the latent vectors $X$ and calculate these statistics:

$$\hat{\mu}_C = \frac{1}{N_C}\sum_{i=0}^{N_C} x_{c_i} \tag{1}$$

$$\Sigma = \frac{1}{\sum_c N_c}\sum_{c=0}^{C}\sum_{i=0}^{N_C}\left(x_{c_i} - \hat{\mu}_c\right)^T\left(x_{c_i} - \hat{\mu}_c\right) \tag{2}$$

We can then calculate the relative likelihood that a sample belongs to a given cluster by measuring the squared Mahalanobis distance between the sample and the cluster's mean. This implicitly models the latent distribution of each cluster as a multivariate normal, with all clusters sharing a tied covariance matrix Σ. We take the minimum of these distances as a metric to indicate whether a sample belongs to the total in-distribution, as belonging to any of the clusters implies membership.

$$d = \min_{c\in\{0,1,\dots,C\}} (x-\hat{\mu}_c)^T\Sigma^{-1}(x-\hat{\mu}_c) \quad (3)$$

We classify a sample as OOD if the value of our metric $d$ is greater than some threshold value $\tau$:

$$D = \begin{cases} 1 \; if \; d \geq \tau \\ 0 \; else \end{cases} \quad (4)$$

The threshold can be selected as the value which correctly identifies a certain percentage of true positive samples. In keeping with other works, we test both a set value at a True Positive Rate (TPR) of 95%, as well as a range of values to calculate the Receiver Operating Characteristic (ROC) curve (Section IV).

This method's advantage over the simpler Euclidean distance lies in its ability to account for ellipsoidal clusters, where L2 would assume that the clusters are always spherical in shape. This allows for the metric to more closely fit the data's distribution.

### *B. Using Mahalanobis for Classification*

By employing this method for OOD detection, we have produced the information necessary to also classify ID samples [16]; we simply assign the sample to the closest class cluster, as measured by squared Mahalanobis distance:

$$c = \operatorname*{argmin}_{c\in\{0,1,\dots,C\}} (x-\hat{\mu}_c)^T\Sigma^{-1}(x-\hat{\mu}_c) \quad (5)$$

### *C. Covariance Matrix Inversion Analysis*

We can apply this method directly to large pretrained models (see Section IV), but the latent space's non-specialized feature set may cause issues in practice due to discrepancies between the sample distribution and the overall span of the latent space. Because it measures a continuous value rather than a simple discrete assignment, OOD detection may be especially sensitive to any noise introduced by this potential misalignment. To understand and address this gap without finetuning or hyperparameter searches, we examine the mechanics of Mahalanobis distance to find the pitfalls of this method when the model has not been trained exclusively on the in-distribution.

Mahalanobis distance is equivalent to the Euclidean distance between projections through a whitening transform, where samples are decorrelated and standardized to a mean of zero and a variance of one; the whitening matrix of Mahalanobis/ZCA whitening [21] for a centered dataset with covariance $\Sigma$ is given as:

$$P_W = \Sigma^{-\frac{1}{2}} \quad (6)$$

In our application, the tied covariance matrix of the samples centered on the class means, as a positive semi-definite matrix, can be inverted via Eigendecomposition:

$$\Sigma = v\Lambda v^T,\ \Sigma^{-1} = v\Lambda^{-1}v^T \quad (7)$$

This yields a whitening matrix of the form:

$$\Sigma^{-\frac{1}{2}} = v\Lambda^{-\frac{1}{2}} \quad (8)$$

Observe that, analogously to z-score normalization in uncorrelated data, the features are divided by the inverse square root of the eigenvalues. Consider the task at hand, where the latent distribution of the training set may span only a small subset of the model's latent space. Many features may be unused, resulting in a lower effective dimensionality and some eigenvalues approaching zero magnitude. Mahalanobis distance would magnify these features, degrading our metric with noise.

To solve this, we employ rank reduction, inspired by Principal Component Analysis (PCA) [22]. PCA decorrelates features, and can reduce the dimensionality of data by truncating a decorrelating projection matrix to exclude low-variance features. To preserve Mahalanobis distance's ellipsoidal clustering while avoiding magnification of irrelevant features, we can truncate our whitening matrix, eliminating features with small eigenvalues and keeping the top-k highest variance projected features:

$$\Sigma_k^{-\frac{1}{2}} = v_k\Lambda_k^{-\frac{1}{2}}, v_k \in \mathbb{R}^{n\times k}, \Lambda_k \in \mathbb{R}^{k\times k} \quad (9)$$

While related to PCA, our method retains dimensions describing the variance around class clusters, rather than the global mean. This distinction ensures that the reduced feature space focuses on features which determine membership or nonconformity w.r.t. the clusters.

Although low rank matrices are noninvertible, we can directly construct a rank-reduced inverse covariance matrix via the eigendecomposition to compute a Reduced-rank Mahalanobis distance (ReMaD):

$$\Sigma_{rr}^{-1} = v_k\Lambda_k^{-1}v_k^T \quad (10)$$

To select the value of $k$, we employ the Explained Variance Ratio method [22] commonly used with PCA, where the number of features to keep is chosen by selecting those which retain a specified ratio $r$ of the data's variance, as measured by the eigenvalues:

$$k(r) = \min_i \frac{\sum_{j=0}^{i} \operatorname{diag}(\Lambda)_j}{\sum_{j=0}^{n} \operatorname{diag}(\Lambda)_j} \geq r\,, r \in (0,1) \quad (11)$$

To select the value of the explained variance ratio threshold without searching over a validation set, we analyze the scree plot of the training data's covariance matrix: if the distribution of the variance is thin-tailed, we set $r = 0.99$ to retain most of the data's descriptive variance while eliminating low-variance features. However, if the distribution is thick-tailed, such a high threshold ratio may result in keeping many noisy features, so we truncate earlier at $r = 0.95$. We quantify this intuition with a simple metric: we truncate early if the smallest eigenvalue at $r = 0.95$ (in the tail) exceeds 1% of the largest eigenvalue's magnitude (at the peak):

$$r = \begin{cases} 0.99 \; if \, \Lambda_{k(0.95)}/\Lambda_0 < 0.01 \\ 0.95 \; else \end{cases} \quad (12)$$

This simple approach allows us to quantify the choice of $r$ without introducing hyperparameter tuning over a validation set, which may not be available for many real OOD detection tasks. It also minimizes both the computational cost and the potential for overfitting.

By substituting $\Sigma_{rr}^{-1}$ into Equations 3 and 5, we obtain a modified distance metric which we can use for both classification and OOD detection:

$$d = \min_{c\in\{0,1,\dots,C\}} (x-\hat{\mu}_c)^T\Sigma_{\mathrm{rr}}^{-1}(x-\hat{\mu}_c) \quad (13)$$

$$c = \operatorname*{argmin}_{c\in\{0,1,\dots,C\}} (x-\hat{\mu}_c)^T\Sigma_{\mathrm{rr}}^{-1}(x-\hat{\mu}_c) \quad (14)$$

This effectively specializes the metric to the target distribution without finetuning the embedding model, offers the advantages of ellipsoidal clustering, and discards noisy features which may hinder proper detection of out-of-distribution

**Algorithm 1** OOD Detection and Classification using Rank-Reduced Mahalanobis Distance

Training data latent embeddings $X$, labels $Y \in [0, 1, \dots, C]$
Test data latent embeddings $Z$
Detection threshold $\tau$
**for** $c \in [0, 1, \dots, C]$:
  Set $N_c = |\{y_i = c\}|$
  Set $X_c = \{x_i \; \forall \; i : y_i = c\}$
  Calculate $\hat{\mu}_c = \frac{1}{N_c} \sum_{i=0}^{N_c} x_{c_i}$
**end for**
Calculate $\Sigma = \frac{1}{\sum_c N_c} \sum_{c=0}^{C} \sum_{i=0}^{N_c} \left(x_{c_i} - \hat{\mu}_c\right)^T \left(x_{c_i} - \hat{\mu}_c\right)$
Perform Eigendecomposition: $\Sigma = v \Lambda v^T$
Calculate $EVR = \frac{\mathrm{diag}(\Lambda)}{\sum_{i=0}^{n} \mathrm{diag}(\Lambda)_i}$
Calculate $k(0.95) = \min_i \frac{\sum_{j=0}^{i} \mathrm{diag}(\Lambda)_j}{\sum_{j=0}^{n} \mathrm{diag}(\Lambda)_j} \geq 0.95$
Select $r = \begin{cases} 0.99 \; if \, \Lambda_{k(0.95)}/\Lambda_0 < 0.01 \\ 0.95 \; else \end{cases}$
Set $k = \min_i \frac{\sum_{j=0}^{i} \mathrm{diag}(\Lambda)_j}{\sum_{j=0}^{n} \mathrm{diag}(\Lambda)_j} \geq r$
Calculate $\Sigma_{rr}^{-1} = v_k \Lambda_k^{-1} v_k^T$
**for** $z_i \in Z$:
  $c_i = \underset{c \in \{0,1,\dots,C\}}{\mathrm{argmin}} \, (z_i - \hat{\mu}_c)^T \Sigma_{\mathrm{rr}}^{-1} (z_i - \hat{\mu}_c)$
  $d_i = \min_{c \in \{0,1,\dots,C\}} (z_i - \hat{\mu}_c)^T \Sigma_{\mathrm{rr}}^{-1} (z_i - \hat{\mu}_c)$
  $D_i = \begin{cases} 1 \; if \; d_i \geq \tau \\ 0 \; else \end{cases}$
**end for**

samples. The full process for preprocessing and inference is detailed in Algorithm 1.

## IV. Experiments

### A. Datasets and Models

We test our method on four target classification datasets: CIFAR-10 [23], CIFAR-100 [23], ImageNet1k [24], and MIDOG [25]. For CIFAR-10, CIFAR-100, and ImageNet1k, we use the testing data splits introduced in OpenOODv1.5 [26] [27], an out-of-distribution detection benchmark which provides standardized testing splits across a large set of benchmark image datasets (Table I). To assess a case outside of most general pretrained models' knowledgebases, we employ the in- and out-of-distribution data splits for the MIDOG dataset introduced by OpenMIBOOD [28], an extension of OpenOOD for OOD detection in biomedical images (Table I). The ID samples include three labeled classes from breast carcinoma imaging: mitotic cells, imposter cells, and background tissue;

TABLE I
Out-of-Distribution Test Sets from OpenOODv1.5 and OpenMIBOOD

| ID | Near-OOD | Far-OOD |
|---|---|---|
| ImageNet1k | SSB-hard [32], NINCO [31] | iNaturalist [33], Textures [34], OpenImages-O [35] |
| CIFAR-10 | CIFAR-100, Tiny ImageNet [36] | MNIST [37], SVHN [38], Textures, Places365 [39] |
| CIFAR-100 | CIFAR-10, Tiny ImageNet | MNIST, SVHN, Textures, Places365 |
| MIDOG 1a | MIDOG 2, 3, 4, 5, 6a, 6b, 7 | CCAgT [41], FNAC2019 [40] |

this represents a difficult task which falls far outside the regularly tested range of zero-shot classification tasks on CLIP models. Both benchmarks divide out-of-distribution testing sets into two broad categories: Near-OOD contains image sets that are similar to the in-distribution, while Far-OOD includes less similar sets.

We employ three widely available large pretrained CLIP models: the original CLIP from OpenAI [3], and the more recent and performant SigLip2 [29] and MetaCLIP2 [30]; we use the "base" version of each of these models with a 16px patch size (commonly referred to as ViT-B/16), offering a balance between capability and size. We measure these model's zero-shot classification capabilities as a baseline for training-free performance by a simplified and data-agnostic method, comparing test images to a set of captions formatted as "a photo of a {class label}." This also allows us to verify our assumption that MIDOG lies outside the knowledge of these models; the accuracy of this method ranged from 22-32% (Table II). For OOD detection, we compare our method to two latent distance-based baselines: Euclidean distance, and unmodified Mahalanobis distance.

### B. Results

Across the three CLIP image embedding models, our method's average performance in Near- and Far-OOD task categories exceeded both baselines in 22/24 cases (Table III). Our rank-reduced Mahalanobis OOD metric consistently achieved higher performance in OOD detection, as measured by Area Under the Receiver Operating Characteristic curve (AUC) and false positive rate at 95% true positive rate (FPR). In classification, our method greatly exceeded the zero-shot text embedding baseline; these methods are not closely comparable due to the baseline's use of the text encoder and our method's use of the training data, but these results provide some insight into the limitations of text label-based classification and the effectiveness of our method at domain transfer into unknown tasks.

TABLE II
Out-of-Distribution Detection Results

| | ImageNet1k | | CIFAR-10 | | CIFAR-100 | | MIDOG | |
|---|---|---|---|---|---|---|---|---|
| | Text Emb. | Ours | Text Emb. | Ours | Text Emb. | Ours | Text Emb. | Ours |
| OpenAI CLIP | 0.5988 | **0.7257** | 0.8820 | **0.9450** | 0.5943 | **0.7723** | 0.3147 | **0.8088** |
| SigLIP2 | 0.6714 | **0.7915** | 0.9394 | **0.9576** | 0.7099 | **0.7942** | 0.2191 | **0.8406** |
| MetaCLIP2 | 0.6623 | **0.7421** | 0.9414 | **0.9723** | 0.7107 | **0.8374** | 0.3028 | **0.8088** |

TABLE III
OUT-OF-DISTRIBUTION DETECTION RESULTS

| | | ImageNet1k Results | | | | | | CIFAR-10 Results | | | | | |
|---|---|---|---|---|---|---|---|---|---|---|---|---|---|
| | | L2 | | Mahalanobis | | Ours | | L2 | | Mahalanobis | | Ours | |
| | | AUC | FPR | AUC | FPR | AUC | FPR | AUC | FPR | AUC | FPR | AUC | FPR |
| OpenAI CLIP | Near | 0.5991 | 0.8026 | 0.6402 | 0.8023 | **0.6575** | 0.8020 | 0.7583 | 0.6418 | 0.9144 | 0.3549 | **0.9209** | 0.3448 |
| | Far | 0.4995 | 0.8610 | 0.7735 | 0.4861 | **0.8249** | 0.4513 | 0.8519 | 0.2924 | 0.9965 | 0.0124 | **0.9972** | 0.0123 |
| SigLip2 | Near | 0.5626 | 0.8485 | 0.7055 | 0.7406 | **0.7362** | 0.7350 | 0.8181 | 0.6393 | 0.9287 | 0.3651 | **0.9316** | 0.3527 |
| | Far | 0.6519 | 0.6445 | 0.8557 | 0.3121 | **0.9051** | 0.2688 | 0.8505 | 0.3168 | 0.9924 | 0.0290 | **0.9939** | 0.0259 |
| MetaCLIP2 | Near | 0.6981 | 0.7315 | 0.7393 | 0.6728 | **0.7594** | 0.6754 | 0.8402 | 0.4379 | 0.9226 | 0.3348 | **0.9320** | 0.3176 |
| | Far | 0.7804 | 0.5356 | 0.8283 | 0.3667 | **0.9064** | 0.2956 | 0.9673 | 0.1130 | 0.9975 | 0.0055 | **0.9994** | 0.0019 |

| | | CIFAR-100 Results | | | | | | MIDOG Results | | | | | |
|---|---|---|---|---|---|---|---|---|---|---|---|---|---|
| | | L2 | | Mahalanobis | | Ours | | L2 | | Mahalanobis | | Ours | |
| | | AUC | FPR | AUC | FPR | AUC | FPR | AUC | FPR | AUC | FPR | AUC | FPR |
| OpenAI CLIP | Near | 0.7310 | 0.7494 | 0.8591 | 0.4156 | **0.8668** | 0.4044 | 0.6003 | 0.8668 | 0.8812 | 0.3011 | **0.8872** | 0.2965 |
| | Far | 0.7460 | 0.4627 | **0.9860** | 0.0576 | 0.9829 | 0.0713 | 0.7491 | 0.4522 | 0.9996 | 0.0020 | **1.0000** | 0.0000 |
| SigLip2 | Near | 0.6987 | 0.7923 | 0.8674 | 0.4424 | **0.8737** | 0.4199 | 0.7583 | 0.6096 | 0.8075 | 0.4491 | **0.8177** | 0.4297 |
| | Far | 0.6458 | 0.5331 | 0.9544 | 0.1482 | **0.9560** | 0.1379 | 0.5739 | 0.8904 | 0.9942 | 0.0159 | **0.9962** | 0.0080 |
| MetaCLIP2 | Near | 0.7668 | 0.5199 | 0.8402 | 0.4073 | **0.8480** | 0.3971 | 0.7472 | 0.6022 | 0.8780 | 0.2783 | **0.8809** | 0.2851 |
| | Far | 0.9017 | 0.2330 | 0.9949 | 0.0114 | **0.9974** | 0.0103 | 1.0000 | 0.0000 | 1.0000 | 0.0000 | 1.0000 | 0.0000 |

## V. ANALYSIS

These results demonstrate that our method can effectively utilize large pretrained image embedding models for new classification and out-of-distribution detection tasks, refining classifier performance without training to increase robustness to anomalous data. Our rank reduction removed significant numbers of features in every case (Fig. 1), lending credence to our hypothesis that a specific dataset may effectively occupy a lower dimensional span of a model's latent space.

A single inference pass over the training set generates the statistics, requiring far less computational resources and lower device memory than finetuning using backpropagation, and no hyperparameter searches are required. Further, it allows domain adaptation to entirely new classification tasks without relying on any part of the trained model possessing relevant prior knowledge. While data is required, training is not, facilitating the rapid deployment of new classifiers as new tasks arise.

Additionally, because the model's weights are not modified for each new task, only the latent mean vectors and rank-reduced inverse covariance matrix must be stored. For each of the tested CLIP models with latent dimensions of 512-768, storing this data takes between 2 and 6MB per dataset, making this method a storage-efficient way to deploy classifiers for many different tasks by using the same pretrained model for all of them.

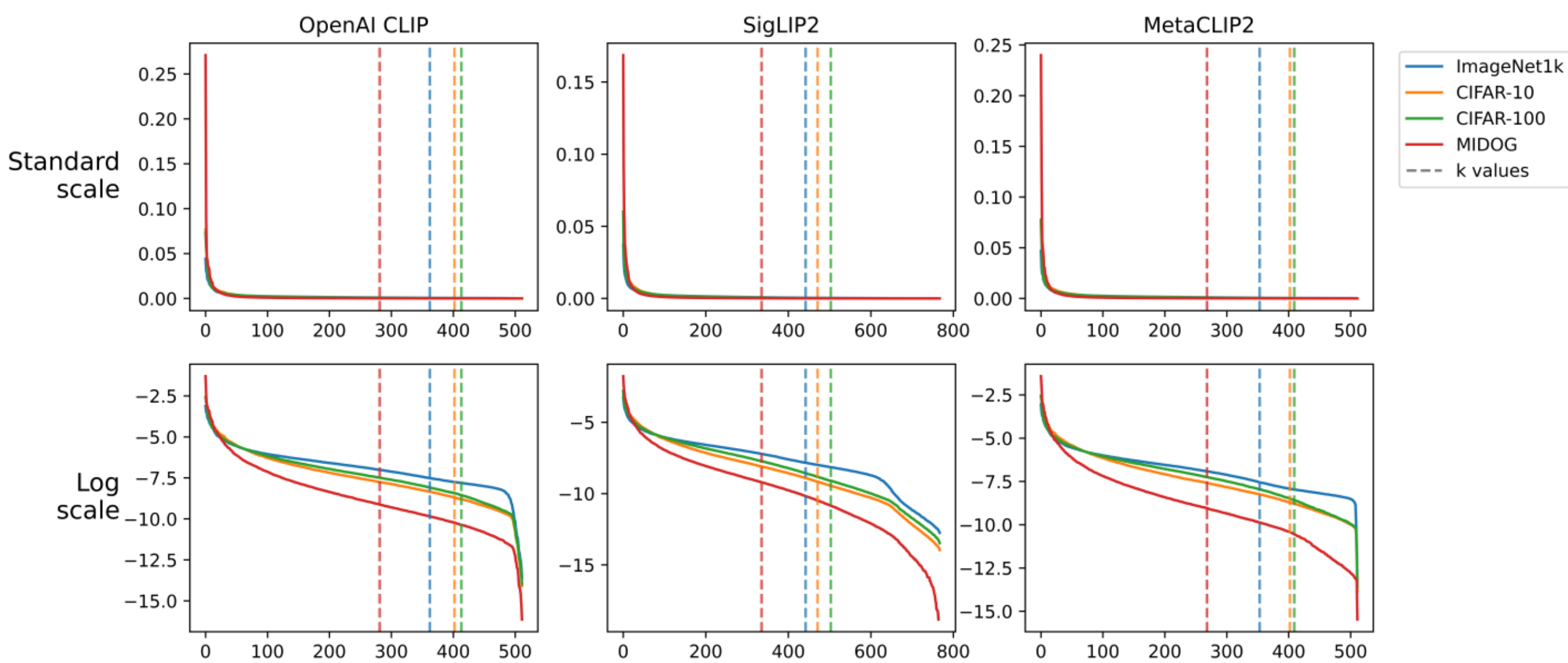

Fig. 1. Normalized scree plots of each dataset's latent space distribution with computed rank thresholds $k$. Log scale plots are included for clarity to illustrate the differences in variance distribution among the datasets.

The observed results provide insight into the application of large pretrained models to new tasks, substantiating the intuition that any specific task uses a measurable subset of the model's latent space features. Further, they indicate that the removal of unused features can increase performance, demonstrating the value of a tuning-free approach to latent space reduction.

## VI. Conclusion

We introduce ReMaD, a novel refinement method for domain adaptation of large pretrained models to new classification domains using only training data, without finetuning through backpropagation or relying on multimodal prior knowledge. We demonstrate competitive performance across four classification tasks, with high robustness to anomalous data as measured by OOD detection scores on a broad spectrum of benchmark datasets. In future work, our method could potentially be augmented to handle few-shot settings without introducing any tuning requirements; this would further expand its applicability to data-restricted tasks while retaining its ability to refine latent distributions for new domains with a single inference pass.